\documentclass[10pt,twocolumn,letterpaper]{article}

\usepackage[pagenumbers]{cvpr} 

\usepackage[dvipsnames]{xcolor}
\usepackage{graphicx}
\usepackage{multirow}
\usepackage{tabularx}
\usepackage{wrapfig}
\usepackage[acronym,toc]{glossaries}
\usepackage[flushleft]{threeparttable}

\definecolor{cvprblue}{rgb}{0.21,0.49,0.74}
\usepackage[pagebackref,breaklinks,colorlinks,citecolor=cvprblue]{hyperref}

\def\confName{CVPR}
\def\confYear{2024}

\title{Team RAS in 9th ABAW Competition: Multimodal Compound Expression Recognition Approach}

\author{Elena Ryumina\\
St. Petersburg Federal Research Center \\ of the Russian Academy of Sciences\\
St. Petersburg, Russia\\
{\tt\small ryumina.e@iias.spb.su}
\and
Maxim Markitantov\\
St. Petersburg Federal Research Center \\ of the Russian Academy of Sciences\\
St. Petersburg, Russia\\
{\tt\small markitantov.m@iias.spb.su}
\and
Alexandr Axyonov\\
St. Petersburg Federal Research Center \\ of the Russian Academy of Sciences\\
St. Petersburg, Russia\\
{\tt\small axyonov.a@iias.spb.su}
\and
Dmitry Ryumin\\
St. Petersburg Federal Research Center \\ of the Russian Academy of Sciences\\
St. Petersburg, Russia\\
{\tt\small ryumin.d@iias.spb.su}
\and
Mikhail Dolgushin\\
St. Petersburg Federal Research Center \\ of the Russian Academy of Sciences\\
St. Petersburg, Russia\\
{\tt\small dolgushin.m@iias.spb.su}
\and
Alexey Karpov\\
St. Petersburg Federal Research Center \\ of the Russian Academy of Sciences;\\
ITMO University \\
St. Petersburg, Russia\\
{\tt\small karpov@iias.spb.su}
}

\begin{document}
\newacronym{CER}{CER}{Compound Expression Recognition}
\newacronym{CE}{CE}{Compound Expressions}
\newacronym{ABAW}{ABAW}{Affective Behavior Analysis in-the-Wild}
\newacronym{AFEW}{AFEW}{Acted Facial Expressions in The Wild}

\newacronym{VS}{VS}{Static visual model}
\newacronym{VD}{VD}{Dynamic visual model}
\newacronym{FCL}{FCL}{Fully Connected Layer}
\newacronym{FPS}{FPS}{Frame Per Second}
\newacronym{LSTM}{LSTM}{Long Short-Term Memory}
\newacronym{BiLSTM}{BiLSTM}{Bidirectional Long Short-Term Memory}
\newacronym{VAD}{VAD}{Voice Activity Detection}
\newacronym{NLL}{NLL}{Negative Log-Likelihood}
\newacronym{SOTA}{SOTA}{State-of-the-Art}

\newacronym{Ne}{Ne}{Neutral}
\newacronym{An}{An}{Anger}
\newacronym{Di}{Di}{Disgust}
\newacronym{Fe}{Fe}{Fear}
\newacronym{Ha}{Ha}{Happiness}
\newacronym{Sa}{Sa}{Sadness}
\newacronym{Su}{Su}{Surprise}
\newacronym{ASR}{ASR}{Automatic Speech Recognition}

\newacronym{RAF-DB}{RAF-DB}{Real-world Affective Faces Database}
\newacronym{CLIP}{CLIP}{Contrastive Language-Image Pretraining}
\newacronym{Jina}{Jina}{Jina Embeddings V3}
\newacronym{RoBERTa}{RoBERTa}{Emotion English DistilRoBERTa Base}
\newacronym{MHPF}{MHPF}{Multi-Head Probability Fusion}
\newacronym{PF}{PF}{Probability Fusion}
\newacronym{LLM}{LLM}{Large Language Model}

\newacronym{PPA}{PPA}{Pair-Wise Probability Aggregation}
\newacronym{PFSA}{PFSA}{Pair-Wise Feature Similarity Aggregation}

\maketitle
\glsresetall
\begin{abstract}
\gls{CER}, a subfield of affective computing, aims to detect complex emotional states formed by combinations of basic emotions. In this work, we present a novel zero-shot multimodal approach for \gls{CER} that combines six heterogeneous modalities into a single pipeline: static and dynamic facial expressions, scene and label matching, scene context, audio, and text. Unlike previous approaches relying on task-specific training data, our approach uses zero-shot components, including \gls{CLIP}-based label matching and Qwen-VL for semantic scene understanding. We further introduce a \gls{MHPF} module that dynamically weights modality-specific predictions, followed by a \gls{CE} transformation module that uses \gls{PPA} and \gls{PFSA} methods to produce interpretable compound emotion outputs. Evaluated under multi-corpus training, the proposed approach shows F1 scores of 46.95\% on AffWild2, 49.02\% on \gls{AFEW}, and 34.85\% on C-EXPR-DB via zero-shot testing, which is comparable to the results of supervised approaches trained on target data. This demonstrates the effectiveness of the proposed approach for capturing \gls{CE} without domain adaptation. The source code is publicly available\footnote{https://github.com/SMIL-SPCRAS/ICCVW\_25}.
\end{abstract}

\glsresetall

\section{Introduction}
\label{sec:intro}

\gls{CER}, as a subfield of affective computing, represents an emerging challenge in intelligent human-computer interaction and multimodal interface design. Unlike traditional emotion recognition systems that focus on discrete, basic emotional states -- such as Fear, Happiness, Sadness, Anger, Surprise, and Disgust -- \gls{CER} aims at detecting complex, blended emotional expressions that naturally arise in human behavior. These \glspl{CE}, including combinations like Happily Surprised, Angrily Surprised, or Sadly Fearful, reflect more realistic and nuanced affective states experienced in everyday interactions.

While significant progress has been made over the past two decades in recognizing prototypical emotions~\cite{RYUMINA2022435,kollias2023multi}, most approaches still fall short in modeling the full spectrum of emotional complexity. Human affect is rarely confined to a single elementary emotion; rather, it often manifests as overlapping or sequential blends of multiple affective components. This calls for advanced models capable of capturing such subtleties through richer feature representations and more sophisticated fusion strategies.

Our approach introduces a novel multimodal framework specifically designed for \gls{CER}. It integrates six diverse modalities -- static and dynamic facial expressions, audio signals, scene context, textual descriptions, scene-label alignment, probability and feature fusion -- into a unified system. In contrast to prior work that relies primarily on static facial features~\cite{richet2025textualized}, our method incorporates temporal dynamics using Transformer- and Mamba-based architectures, enabling more accurate modeling of affective evolution over time.

Moreover, unlike systems that require extensive fine-tuning on labeled \gls{CE} corpora~\cite{liu2025abaw7,liu2025compound}, our pipeline leverages zero-shot-capable components. These include \gls{CLIP}-driven label matching and semantic scene analysis using Qwen-VL, allowing the model to generalize beyond predefined emotion classes without domain-specific retraining. We also propose a \gls{MHPF} model that adaptively balances contributions from individual modality-specific predictors.

Finally, we introduce a compound emotion transformation module that combines distributional and similarity-based techniques to project model outputs into interpretable compound emotion categories. This dual-strategy approach enhances both the flexibility and explainability of our system, setting it apart from \gls{CER} pipelines.

\section{Related Work}
\label{sec:rw}

We compare our proposed method with the \gls{SOTA} methods presented in the scope of the 7th, 8th \gls{ABAW} Competitions~\cite{kollias20246th}. This challenge has seen the development of several \gls{CER} systems, including our own~\cite{ryumina2024zero}.

Savchenko~\cite{savchenko2025smoothing} proposed a visual-only method using lightweight EmotiNet models pre-trained for basic facial expression recognition. \gls{CE} predictions were obtained by aggregating the outputs of basic emotions and applying temporal smoothing with box and Gaussian filters. This method enhanced stability while maintaining computational efficiency, making it applicable to real-time scenarios.

Liu et al.~\cite{liu2025compound} proposed a purely visual method for \gls{CER} that employs a curriculum learning strategy. The model is first trained to recognize basic emotions using large-scale corpora containing single-expression samples. To simulate compound expressions, the authors generate hybrid samples using CutMix and Mixup techniques. Compound data is introduced progressively across multiple training stages, with increasing task complexity. A Masked Autoencoder serves as the feature extractor and is fine-tuned on diverse facial corpora. This staged training method is designed to improve the robustness and generalization of the model to compound facial expressions.

Richet et al.~\cite{richet2025textualized} proposed and evaluated two multimodal methods for \gls{CER}: one feature-based and the other text-based. The feature-based method employed modality-specific backbones, ResNet50 for visual data, VGGish for audio, and BERT for text, integrated via co-attention fusion. In contrast, the text-based method converted visual and audio signals into textual descriptions, which, along with transcripts, were processed by the LLaMA-3 model. Experimental results on the C-EXPR-DB and MELD corpora demonstrated that the text-based method performed well when rich transcripts were available (e.g., in MELD), but its performance deteriorated under sparse textual input. The feature-based method exhibited greater robustness in real-world scenarios and better preservation of modality-specific information.

Liu et al.~\cite{liu2025abaw7} proposed a visual-only method for \gls{CER} that combines convolutional and transformer-based models. Local facial features were extracted using ResNet50, while global representations were derived from a Vision Transformer (ViT). In addition, facial action units (FAUs) were extracted using OpenFace and used as handcrafted features. To align features of different dimensionalities, an affine transformation module was applied prior to temporal modeling with a transformer encoder. Final predictions were generated by fusing the outputs of all components through a multilayer perceptron, effectively combining local and global facial expressions for robust compound emotion recognition.

In the methods proposed by~\cite{liu2025abaw7, liu2025compound}, additional corpora annotated with \glspl{CE} were used to train models. Therefore, our method is more comparable to those of~\cite{richet2025textualized, savchenko2025smoothing}, which rely on pretrained components and avoid direct training on the target corpus, thus operating in a zero-shot or lightly adapted setting.
\section{Proposed Method}
\label{sec:method}
\begin{figure*}
  \centering
   \includegraphics[width=0.95\linewidth]{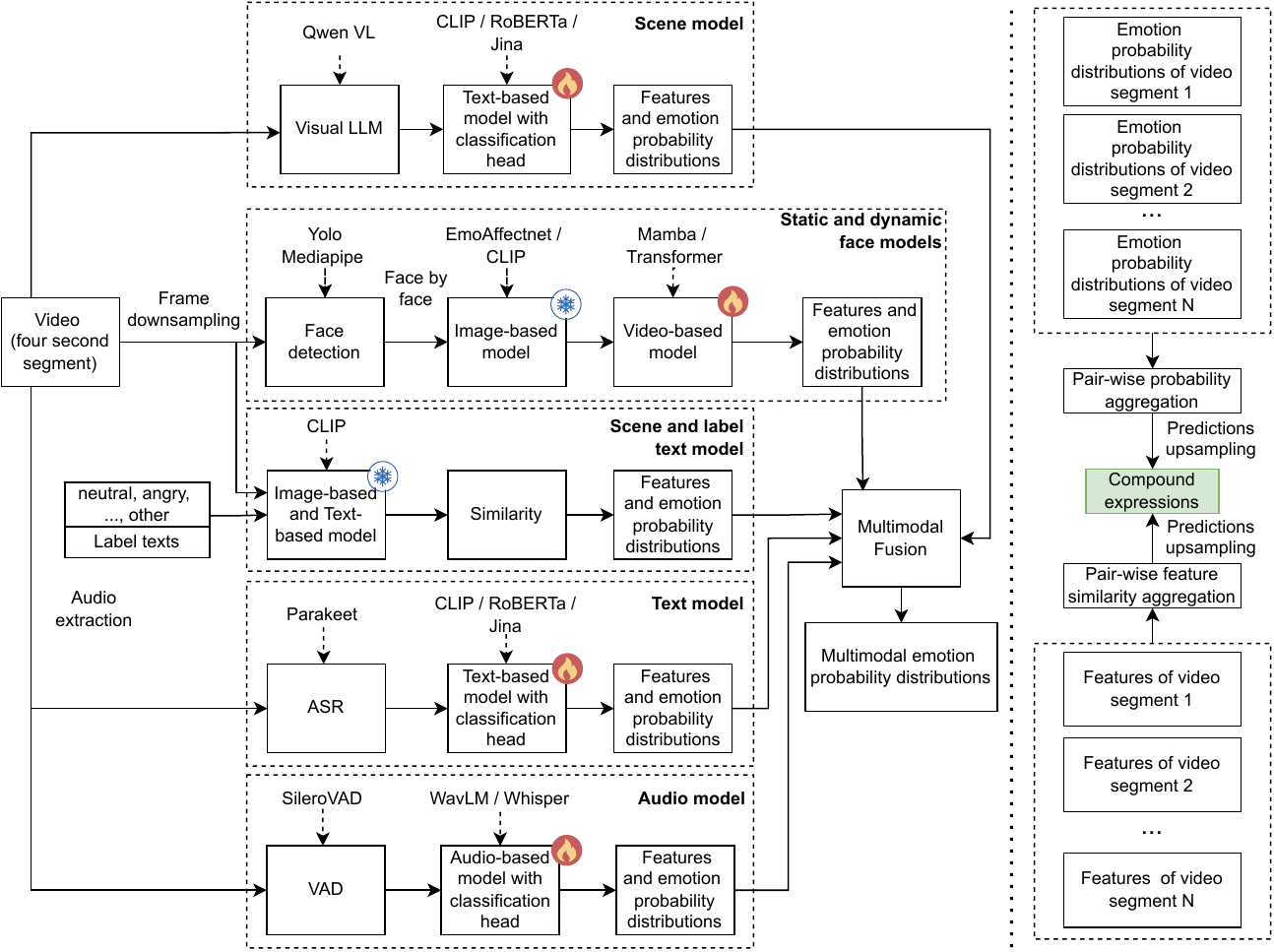}
   \caption{Pipeline of the proposed multimodal \gls{CER} approach.}
   \label{fig:pipeline}
\end{figure*}

The pipeline of the approach proposed is presented in Figure~\ref{fig:pipeline}. At the initial stage, the video files are divided into 4-second segments with a 2-second overlap, since previous studies have shown that this window size is sufficient to extract multimodal emotional patterns. The proposed approach combines six different modalities, which are detailed in the following.

\subsection{Face Models}
For each segment, the frame rate is downsampled to 20 frames per 4-second window, ensuring uniform temporal resolution and reducing computational complexity. Facial detection is performed using a two-stage method: an initial YOLO v11\footnote{https://github.com/akanametov/yolo-face} model, pre-trained on WIDER FACE, and fine-tuned for real-world conditions, detects faces under challenging conditions, while a secondary refinement step based on MediaPipe FaceMesh~\cite{lugaresi2019mediapipe} validates face geometry and filters out false positives detections.

To extract static facial features, two models are used. EmoAffectNet~\cite{RYUMINA2022435}, a ResNet-50-based model fine-tuned on AffectNet, classifies six basic emotions (anger, disgust, fear, happiness, sadness, surprise) and neutral states. A \gls{CLIP}~\cite{radford2021learning} model, based on ViT-B/32 with 12 layers and 12 attention heads. This model was optimized for cross-modal matching between images and text, and simultaneously extracts both textual and visual features.

Temporal modeling is carried out using the Mamba~\cite{gu2023mamba} and Transformer~\cite{vaswani2017attention} models. The Mamba uses a selective state-space mechanism that efficiently models long sequences with linear complexity. The Transformer captures global dependencies using self-attention, but at a higher computational cost.

\subsection{Scene and Label Text Model}
We use \gls{CLIP} to generate bimodal features for video frames and eight emotion labels.  
We compute the cosine proximity between frame and label features to quantify their matching, and then apply softmax normalization for probability predictions. 
By matching visual and textual features, the model supports zero-shot emotion classification and can be extended to other multimodal tasks.

\subsection{Scene Models}
To include contextual semantics, an additional modality is introduced based on textual descriptions of a scene. For each 4-second video segment, a textual description is generated using the vision-language model Qwen-VL 2.5 32B\footnote{https://huggingface.co/Qwen/Qwen-VL-Chat-Int4}, which processes visual data and produces unlimited natural language output. Two prompting strategies are considered. The first prompt encouraged a free-form emotional interpretation of a person’s non-verbal behavior and the surrounding environment, focusing on the dynamics of facial expressions, gestures, and posture. The second prompt constrained the output to include exactly one of a predefined set of eight basic or seven compound emotion labels, either embedded within the description or added as a final classification. 

The generated descriptions, aligned with compound emotion labels via corpus annotations, are encoded using three transformer-based models: \gls{CLIP}~\cite{radford2021learning}, \gls{RoBERTa}\footnote{https://huggingface.co/j-hartmann/emotion-english-distilroberta-base/}, and \gls{Jina}~\cite{sturua2024jina}. A linear classification head was added to each encoder. 

\subsection{Audio Models}
From each video segment, an audio is extracted and passed through Silero\footnote{https://github.com/snakers4/silero-vad} \gls{VAD}. Only segments in which speech was detected are used for training to avoid the influence of non-speech segments on model training. We propose three models for speech emotion recognition.

The first model employs a pre-trained WavLM model~\citep{chen2022wavlm} as a feature extractor, fine-tuning only the upper encoder and projection layers. To capture local temporal dependencies in audio, a Residual \gls{BiLSTM} Block is used. An Attention Pooling then aggregates the sequence information into fixed-size feature vector. The final classification head consists of several \glspl{FCL} with ReLU, LayerNorm, and Dropout, and predicts eight emotion labels.

Unlike the first model, which produces deterministic results, the second model provides a distribution-based prediction with explicit confidence estimation. This model is based on the WavLM model~\citep{chen2022wavlm} and consists of a Temporal Relation block that applies multi-head attention and combines temporal information with an average pooling. These features are concatenated, processed through \glspl{FCL} and passed to the Emotion Uncertainty head, which predicts class-wise means and log-variances. 

\subsection{Text Models}
For text analysis, transcriptions are extracted from 4-seconds audio segments using the Parakeet Token-and-Duration Transducer 0.6B V2 (Parakeet) model\footnote{https://huggingface.co/nvidia/parakeet-tdt-0.6b-v2} designed for English transcription. The original attention mechanism is replaced with a local attention block using a $128\times128$ relative position window to improve computational efficiency on longer samples at the cost of transcription accuracy.

Several transformer-based models are used to extract textual features: \gls{CLIP}~\cite{radford2021learning}, \gls{RoBERTa}\footnote{https://huggingface.co/j-hartmann/emotion-english-distilroberta-base/} and \gls{Jina}~\cite{sturua2024jina}.

Due to the short length of most segments, only the first 1024 tokens of \gls{Jina} features are used. For other models, the maximum supported token length is used. The resulting features are then pooled and passed to the final classification head, consisting of a Dropout layer with a \gls{FCL} to predict eight emotion labels.

\subsection{Modality Fusion Models}
We consider the fusion of modalities at the probability level and at the feature level. The fusion of multi-modal predictions is performed at the probability output level through a learnable convex combination mechanism. Given input probability distributions from multiple models, the \gls{MHPF} model applies a multi-head weighted averaging strategy, where each head independently learns class-specific weights to combine input distributions across modalities. These per-head outputs are then aggregated using an additional set of learnable coefficients, enabling dynamic balancing of head contributions based on class-specific confidence patterns. 

\subsection{Compound Emotion Mapping}
To convert the original eight emotion labels to a set of seven compound emotions, we propose two methods: \gls{PPA} and \gls{PFSA}.

The \gls{PPA} method is based on summarizing the predicted probabilities of the corresponding basic emotions to estimate the probabilities of compound emotions, assuming that compound emotions can be represented as additive combinations of their corresponding basic emotions.

The \gls{PFSA} method is based on cosine proximity in the model’s latent space. First, emotion-specific features are computed from the validation subsets by averaging feature vectors of correctly classified samples in each basic emotion. These features serve as prototypes for basic emotions. The compound emotion prototypes are then constructed by averaging the features of their corresponding basic emotions, followed by L2 normalization. Each feature vector from the test subset is compared with all compound emotion prototypes using cosine similarity. The resulting similarity scores are propagated through a softmax function with temperature scaling to generate a smooth distribution over the compound emotions while preserving sharpness in confident cases.

To produce stable and consistent frame-wise \gls{CER}, segment-level predictions are duplicated across the corresponding frames. Overlapping predictions are averaged per frame to maintain temporal coherence in the final \gls{CER}.

\section{Experiments}

\begin{table*}[t]
\centering
\resizebox{\textwidth}{!}{
\begin{tabular}{@{}lcccccccccc@{}}
\toprule
\multirow{3}{*}{ID}&\multirow{3}{*}{Model} & \multicolumn{6}{c}{Test corpus} \\
&&\multicolumn{3}{c}{AffWild2 (8cl)} & \multicolumn{3}{c}{AFEW (8cl)} & \multicolumn{2}{c}{C-EXPR-DB (7cl), macro-F1} \\
&& macro-F1 & UAR& Average &macro-F1 & UAR & Average& \gls*{PPA} & \gls*{PFSA} \\
\midrule
1& Face \& EmoAffectnet & 26.08 & 42.08 & 34.08 & 45.90 & 45.69 & 45.80 & -- &  --\\
2& Face \& EmoAffectnet \& Transformer & 38.66 & 42.83 & 40.74 & 37.52 & 36.49 & 37.01& -- & -- \\
3& Face \& EmoAffectnet \& Mamba & 35.96 & 40.61 &38.29 & 38.26 & 37.70 &37.98& -- &  --\\
4& Face \& CLIP \& Transformer & 41.53 & 43.40 &42.47& 34.23 & 34.20 & 34.22&--  & -- \\
5& Face \& CLIP \& Mamba & 40.49 & 45.47 &42.98& 37.60  & 37.31& 37.46& 16.79 & -- \\
6& Scene \& LLM \& Jina & 34.60 & 43.33 & 38.97 & 31.48 & 31.70 & 31.59 &--& --\\
7& Scene \& LLM \& Roberta  & 36.31 & 44.68 & 40.49 & 32.30 & 32.04 & 32.17 &34.62& \textbf{34.85}\\
8& Scene \& LLM \& CLIP  & 29.01 & 39.29 & 34.15 & 24.38 & 25.38 & 24.88 &--&--\\
9& Audio \& WavLM \& U-aware & 33.45 & 33.88 & 33.67 & 26.44 & 26.33 & 26.39 &--&-- \\ 
10& Audio \& WavLM \& ReBiLSTM & 26.38 & 32.78 & 29.58 & 23.09 & 23.26 &  23.18 &--& --\\ 
11& Text \& Jina & 32.54 & 41.67 &37.11& 14.36 & 13.90 &14.13&  --&-- \\
12& Text \& Roberta & 27.21 & 35.71& 31.46& 16.24 & 15.40 &15.82& -- &  --\\
13& Text \& CLIP & 35.83 & 45.83 &40.83& 17.15 & 17.40 & 17.25&-- & -- \\
14& Scene-Label text \& CLIP \& Cosine similarity& 10.84 & 13.09 &11.97& 11.79& 30.41& 24.10& --&  --\\
\midrule
\multicolumn{10}{c}{The top three best performing fusion combinations} \\
\midrule
15& Models ID 1 \& 5 \& 6 \& 10 \& 14 \& MHPF &45.21&50.54 & 47.88& 46.41& 46.47 &46.44 &--& --\\
16& Models ID 1 \& 5 \& 6 \& 10 \& 13 \& MHPF & 46.33& 50.62& 48.47 & 44.30& 45.11 & 44.71&--& --\\
17& Models ID 1 \& 5 \& 6 \& 10 \& 13 \& 14 \& MHPF & \textbf{46.50} & \textbf{51.04} & \textbf{48.77} & \textbf{47.14} & \textbf{47.53} & \textbf{47.34}&27.08& --\\
\bottomrule
\end{tabular}
}
\caption{Experimental results. PF is probability fusion. \acrshort*{PPA} is \acrlong*{PPA}, \acrshort*{PFSA} is \acrlong*{PFSA}.}
  \label{tab:Results}
\end{table*}

During training all video-based models, hyperparameters were systematically varied, including hidden state dimensions (128, 256, 512), classification feature vectors (128, 256, 512), layer counts ($1-4$), Mamba-specific parameters (d\_state: $4-16$; kernel\_size: $1-7$), and Transformer attention heads ($2-16$). The training protocol employed a fixed batch size of 64, 100 epochs, a learning rate of $1e-5$, dropout of 0.15, and the Adam optimizer. Early stopping was triggered if validation metrics plateaued for 25 consecutive epochs, ensuring robust convergence.

Scene-based models were trained for 75 epochs with early stopping triggered after 10 validation epochs without improvement. Batch sizes were set to 16 for \gls{RoBERTa} and 48 for both \gls{Jina} and \gls{CLIP}. Maximum token lengths were fixed at 192 for \gls{RoBERTa} and \gls{Jina}, and 77 for \gls{CLIP}. Training was conducted using the Adam optimizer with a learning rate of $1e-5$.

All audio-based models were trained using the AdamW optimizer and a CosineAnnealingLR scheduler over 100 epochs with early stopping based on validation loss stagnation over 10 epochs. The first two models were trained using the Focal loss~\cite{wang2022soft}, which combines focal weighting and class balancing with label smoothing. The last model used \gls{NLL} loss with additional Kullback-Leibler divergence regularization and log-variance penalization. Additionally, this model employed a progressive unfreezing schedule, gradually unlocking deeper layers of the WavLM model to refine feature extraction without destabilizing training.

Text-based models were trained for 100 epochs with early stopping, triggered if validation metrics did not improve for 30 consecutive epochs. Batch sizes of 16, 32, and 64 were used for \gls{Jina}, \gls{RoBERTa}, and \gls{CLIP}, respectively, with the final 4, 3, and 2 layers unfrozen. The training protocol consisted of a learning rate of $1e-4$, and the Adam optimizer.

\begin{table}[t]
\caption{Performance comparison (F1, \%) of the \acrshort*{SOTA} methods. V is video modality, AVT is audio, video, and text modalities. ZH is zero-shot setup. FT is fine-tuned models on target task.
}
\centering
\resizebox{\columnwidth}{!}{
\begin{tabular}{@{}lccc@{}}
\toprule
Method & Setup & Modality & C-EXPR-DB\\
\midrule
Liu et al.~\cite{liu2025abaw7} &FT&V&22.81\\
Richet et al.~\cite{richet2025textualized} &ZH&AVT&25.91\\
Savchenko~\cite{savchenko2025smoothing}&ZH&V&32.43\\
Liu et al.~\cite{liu2025compound}&FT&V&60.63\\
\midrule
Ours&ZH&V&34.86\\
\bottomrule
\end{tabular}
}
  \label{tab:SOTA}
\end{table}

\subsection{Research corpora}

For the purpose of our study, we first train our models for emotion recognition using two corpora: \gls{AFEW} and AffWild2~\cite{kollias2020analysing,kollias2021distribution}. The \gls{AFEW} dataset contains annotations for six basic emotions and a neutral state, whereas AffWild2~\cite{kollias2024distribution,kollias20246th} includes an additional eighth class labeled "other". To enable joint training on both corpora, we introduce a placeholder "other" class for \gls{AFEW} to align its label space with that of AffWild2~\cite{kollias2021affect,kollias2021analysing}.

Both corpora consist of two official splits: a training set (773 videos or 855 4-second segments for \gls{AFEW} and 248 videos or 15158 4-second segments for AffWild2~\cite{kollias2022abaw,kollias2023abaw}) and a validation set (383 videos or 411 4-second segments for \gls{AFEW} and 70 videos or 6450 4-second segments for AffWild2~\cite{kollias2023multi,kollias2023abaw2}). While both corpora capture facial expressions in naturalistic settings, \gls{AFEW} specifically features acted emotional expressions. The domains of both corpora are relatively close to the target corpus, C-EXPR-DB~\cite{kollias2023multi}, for \gls{CER}, for which only 56 videos (449 4-second segments) are available for evaluation.

\subsection{Experimental Results}
The experimental results demonstrate (in Table~\ref{tab:Results}) varying performance across different model configurations on three benchmark datasets: AffWild2 (8 classes), \gls{AFEW} (8 classes), and C-EXPR-DB (7 classes). Individual models utilizing visual features show moderate effectiveness, with combinations of face-based features and EmoAffectNet achieving 34.08 average performance on AffWild2 and 45.80 on \gls{AFEW}. When enhanced with temporal modeling using either Transformer or Mamba architectures, these models exhibit divergent behavior: the Transformer variant improves AffWild2 performance to 40.74 but degrades \gls{AFEW} results to 37.01, while Mamba achieves more balanced gains of 38.29 and 37.98 respectively. Similarly, CLIP-based visual representations combined with Mamba yield competitive scores of 42.98 (AffWild2) and 37.46 (\gls{AFEW}), outperforming their Transformer counterparts.

Models incorporating scene-level features and language encoders present mixed outcomes. Among them, the combination of scene features with \gls{RoBERTa} embeddings achieves 40.49 on AffWild2 and 32.17 on \gls{AFEW}, along with notable performance on C-EXPR-DB~\cite{kollias20247th} (macro-F1 of 34.62 by pair wise sum method and macro-F1 of 34.85 by feature similarity method), suggesting its potential for capturing contextual cues. In contrast, \gls{Jina}-embeddings perform slightly weaker overall but maintain stable recognition rates across both emotion corpora. Audio-based models generally underperform compared to visual and textual modalities, with WavLM and U-Net-aware frontends yielding average scores around 33.67 and 29.58 on AffWild2, respectively. Text-based models, particularly those leveraging \gls{CLIP} embeddings, achieve relatively high AffWild2 scores (up to 40.83), although they significantly drop on \gls{AFEW}, indicating domain mismatch or limited linguistic expressiveness in that dataset.

The best performing multimodal configuration employing \gls{MHPF}  on models ID 1, 5, 6, 10, 13, and 14 achieves the highest scores on AffWild2 (48.77 average two metrics) and \gls{AFEW} (47.34 average). Notably, this ensemble also attains a macro-F1 of 27.08 on C-EXPR-DB~\cite{kollias2025dvd}, demonstrating its generalization capability across domains and annotation schemes. Despite the gain on emotional corpora due to multimodal fusion, on the C-EXPR-DB~\cite{kolliasadvancements} corpus \gls{CER}, the best performance (macro-F1 = 34.85) is achieved by the scene-based model.  This result shows that with correct prompting for a vision-language model, reliable results can be achieved for both basic emotion recognition and \gls{CER}. Comparison with the \gls{SOTA} results (Table~\ref{tab:SOTA}) shows that the proposed approach outperforms all the approaches~\cite{liu2025abaw7,savchenko2025smoothing} that were not trained on the target task, losing only to the approaches~\cite{liu2025compound}, which do not represent a zero-shot solution.
\section{Conclusions}

In this paper, we propose a novel multimodal approach for \gls{CER} that integrates six modalities: static and dynamic facial expressions, audio, scene context, text, scene-label alignment, as well as probabilistic fusion. Each modality is processed using zero-shot or general-purpose models, including \gls{CLIP}, Qwen-VL, WavLM, and RoBERTa, while temporal dynamics are modeled using Transformer and Mamba architectures. A \gls{MHPF} dynamically combines emotion probability distributions, followed by a \gls{CE} transformation step based on distributional and similarity-based matching.

Evaluated under a multi-corpus training setup, the method achieves F1 scores of 46.95\%, 49.02\%, and 34.85\% on AffWild2, AFEW, and C-EXPR-DB, respectively -- without any fine-tuning on target data. The best performance on AffWild2, AFEW  was achieved by a multimodal fusion, while on C-EXPR-DB was achieved by a vision-language model, indicating the potential of prompt-based strategies and scene description in~\gls{CER}. These results are competitive with supervised approaches, demonstrating the effectiveness of our zero-shot pipeline.

{
    \small
    \bibliographystyle{ieeenat_fullname}
    \bibliography{main}
}


\end{document}